\documentclass[letterpaper]{article} 
\usepackage{aaai23}  
\usepackage{times}  
\usepackage{helvet}  
\usepackage{courier}  
\usepackage[hyphens]{url}  
\usepackage{graphicx} 
\usepackage{natbib}  
\usepackage{caption} 
\usepackage{algorithm}
\usepackage{algorithmic}

\usepackage{newfloat}
\usepackage{listings}
\DeclareCaptionStyle{ruled}{labelfont=normalfont,labelsep=colon,strut=off} 
\floatstyle{ruled}
\newfloat{listing}{tb}{lst}{}
\floatname{listing}{Listing}
\title{Planning and Learning: Path-Planning for Autonomous Vehicles, a Review of the Literature}
\author{
 Kevin Osanlou, Christophe Guettier, Tristan Cazenave, Eric Jacopin}
\affiliations{
Safran, kevin.osanlou@safrangroup.com\\
Safran, christophe.guettier@safrangroup.com\\
Paris-Dauhine University, tristan.cazenave@lamsade.dauphine.fr\\ Ecoles Saint-Cyr, eric.jacopin@st-cyr.terre-net.defense.gouv.fr
}

\usepackage{bibentry}

\def\eg{\emph{e.g.~}} 
\def\ie{\emph{i.e.~}}

\def\etal{\emph{et al.~}}

\def\and{\textit{AND }}

\def\etal{\emph{et al.}}
\def\ie{\emph{i.e. }}
\def\eg{\emph{e.g. }}

\newcommand{\real}{\mathbb{R}}

\newcommand{\thetab}{{\boldsymbol \theta}}

\def\xb{\mathbf{x}}
\def\yb{\mathbf{y}}
\makeatother

\usepackage{amsmath,amssymb,array}
\usepackage{algorithm}

\begin{document}

\maketitle

\begin{abstract}
This short review aims to make the reader familiar with state-of-the-art works relating to planning, scheduling and learning. First, we study state-of-the-art planning algorithms. We give a brief introduction of neural networks. Then we explore in more detail \textit{graph neural networks}, a recent variant of neural networks suited for processing graph-structured inputs. We describe briefly the concept of reinforcement learning algorithms and some approaches designed to date. Next, we study some successful approaches combining neural networks for path-planning. Lastly, we focus on temporal planning problems with uncertainty.
\end{abstract}

\section{Planning}
The aim of planning is to conceive plans in order to achieve a particular goal. Those plans represent a sequence of actions executed by an agent that enable the transition from a \textit{start state} of an environment, where goal requirements are not satisfied, to an \textit{end state} where they are. In some planning tasks, states are fully observable, in others, only partially. Actions taken by the agent can be deterministic (\ie lead to a certain future state) or non-deterministic (\ie lead to different future states based on some probabilities that are either known or not). State variables can be continuous or not, resulting in a possibly finite or infinite number of states. Actions can be taken in parallel or only one at a time, and have a duration or not. There can be several initial start states or only one. There can be several agents or only one.
Planning environments can be diverse, varying from simple positioning in a graph to the complex dynamics of a \textit{first person shooter} (FPS) video game. 

In classical planning, models are restricted in the following aspects. The environment is fully observable, there is a single agent, states are finite, there is only one known initial start state, actions are instantaneous and deterministic: there is no uncontrollable event. Actions can only be taken one at a time. Therefore, a sequence of actions from a start state will accurately define the end state, which needs to satisfy goal requirements. Generally, classical planning can be represented mathematically by a set ($S,A,P$) where:

\begin{itemize}
    \item $S$ is the set of states
    \item $A$ is the set of actions
    \item $P$ is a state transition function
    
\end{itemize}

\noindent The state transition function $P: S \times A \longrightarrow 2^S$ defines a transition from a current state $s\in S$ to another state $s' \in S$ by considering an action $a \in A$.

To express and solve planning tasks in computer science, different languages have been proposed. 
Each language represents components of the planning environment differently. These include the Stanford Research Institute 
Problem Solver (STRIPS) \citep{fikes1971strips} from SRI International and the popular Planning Domain Definition Language (PDDL) \index{Languages! PDDL} \citep{mcdermott1998pddl}.


\subsection{Applications of Planning in Autonomous Systems}
A system is considered autonomous if it is able to generate and execute plans to achieve its assigned goals without human intervention, and if it is able to deal with unexpected events. Planning has benefited autonomous systems greatly in the past 50 years. Early on, Shakey the robot \citep{nilsson1984shakey}, the first general-purpose autonomous mobile robot, was a project that saw the rise of a powerful planning algorithm known as A* (introduced in the next sections), still used nowadays. Space exploration has benefited greatly from planning techniques. Autonomy in satellites or other space vehicles reduces the need of human presence as well as communication to ground, which can be especially useful for long term missions. Applications include Deep Space 1 \citep{muscettola1998remote}, or more recently the Curiosity rover \citep{rabideau2017prototyping} which is currently exploring Mars. Aerospace applications include Unmanned Aerial Vehicles (UAVs). UAVs can be used for operational situations such as search and rescue tasks. Search and rescue operations are typically very costly both in terms of costs and human resources, and can present human risks. UAVs reduce those costs and their ability to fly autonomously allows to remove human presence for dangerous tasks. Autonomous Unmanned Ground Vehicles (AUGV) and Autonomous Ground Vehicles (AGV) are also at the center of automation efforts where (trajectory) planning is playing a crucial role. Among AGVs, self-driving cars have been the main focus for civil applications given the potentially revolutionary impact they can have on society. The most advanced self-driving cars combine the latest sensors and computer vision tools for environment perception and use planning to make relevant decisions. We refer the reader to \citep{badue2020self} for a complete survey on self-driving cars. AUGVs on the other hand are intended for other tasks such as typical disaster relief situations, in which they can be required to perform technical actions (\eg observations, measurements, communications, etc...) while navigating mostly in off-road environments across defined trajectories \citep{guettier2016constraint}. Automation allows AUGVs to perform dangerous tasks without human presence, and AGVs to move on their own while passengers can focus on other activities.

\section{Motion Planning and Path-Planning}

\begin{figure}[tbh]
\centering
\includegraphics[scale=0.4]{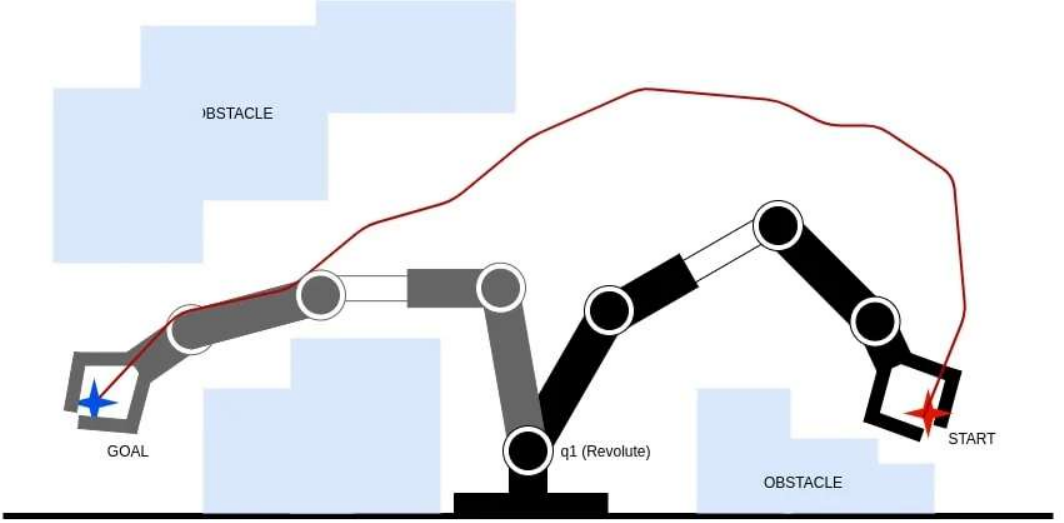}
\caption[A motion planning task for a robot manipulator] {\textbf{A motion planning task for a robot manipulator.} The robot has to carry an object from the START location to the GOAL position. Source: \citep{robotmanip}}
\label{fig_motion_planning}
\end{figure}

Path-planning consists of finding a path leading to a desired point from a start point. Motion planning consists of determining motion and path decisions for an agent in order to allow it to achieve a specified motion-related task. Motion planning is more general than path-planning in the sense that, in addition to 
determining a path the agent needs to take to reach an end point from a start point, it also requires motion characteristics for the agent to reach the end point. Such characteristics can be, but are not limited to, a sequence of positions over time, acceleration values to provide in order to reach a potentially required speed or parameters such as directional angles. Figure ~\ref{fig_motion_planning} illustrates the example of a motion planning task in which a robot manipulator is tasked with grabbing an object located at a START position and moving it to the GOAL position \citep{robotmanip}. The robot has 4 joints which can revolve. The last joint is used to grab and release objects. Let $\alpha_1, \alpha_2, \alpha_3, \alpha_4$ be the angles for each joint, starting from the base of the robot. A planning state is defined by a vector $s = (\alpha_1, \alpha_2, \alpha_3, \alpha_4)$ which entirely defines the position of the robot and the potential object it is carrying. The configuration space $S$ is made of all possible combinations of values each $\alpha_i$ can take. Some states are 'legal', \ie the robot can actually be in those states, others 'illegal', \ie the robot cannot be in those states. For instance, supposing the angle axis is horizontal and revolves counterclockwise, any state written as $(\frac{3}{2}\pi, \alpha_2, \alpha_3, \alpha_4)$ will not be valid regardless of the values of $\alpha_2, \alpha_3, \alpha_4$ since the first joint cannot bend the joined arm downward. By discretizing values of each angle, the robot can determine a series of consecutive angle changes for each joint, which will be considered as actions, that will allow it to grab and move the object from START to GOAL. 


\begin{figure}[tbh]
\centering
\includegraphics[scale=0.3]{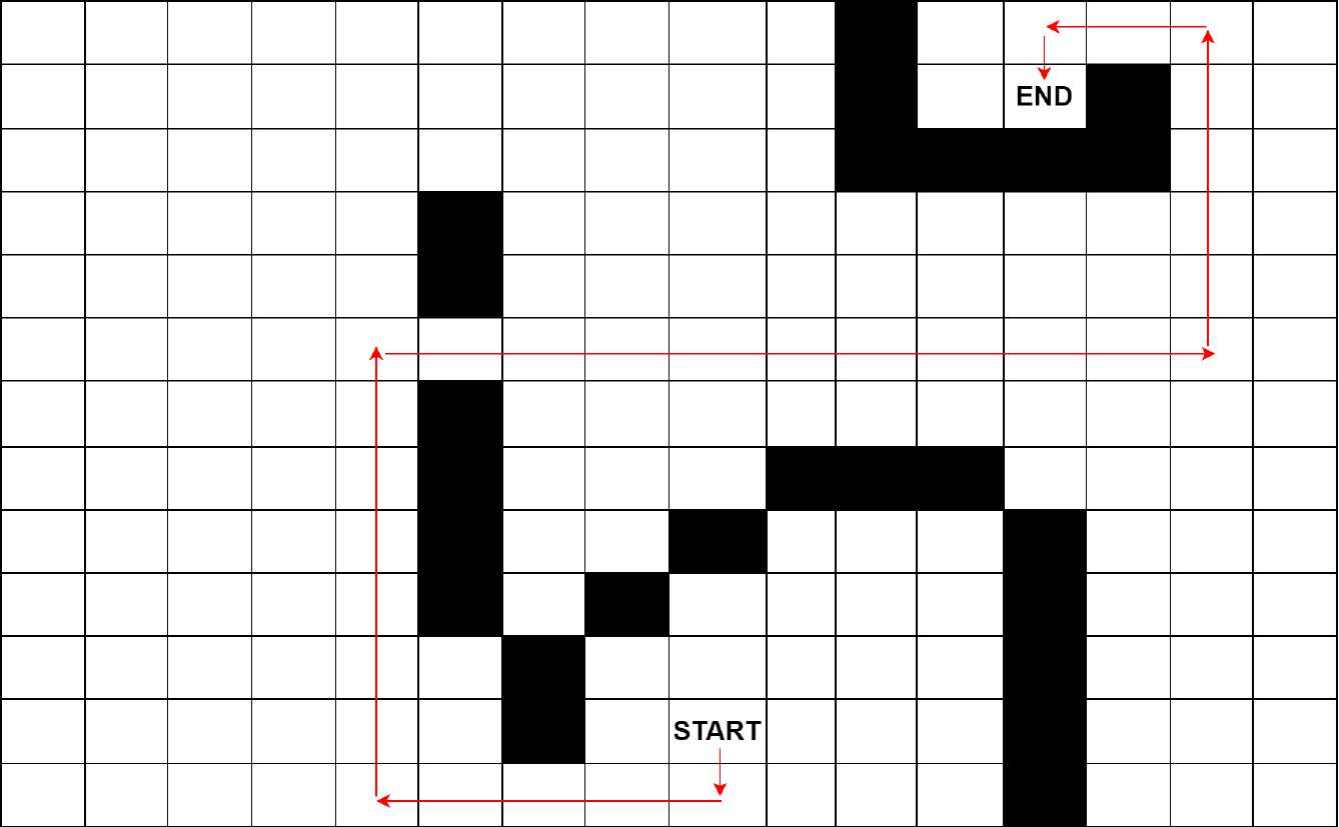}
\caption[A path-planning task for an agent] {\textbf{A path-planning task for an agent.} Black grids represent obstacles. The agent is located at the START grid and needs to move to the END grid. The red arrows represent a possible path to satisfy that goal.}
\label{fig_path_planning}
\end{figure}

Figure~\ref{fig_path_planning} shows a path-planning problem in which an agent has to move from a start grid to an end grid. In this environment, black grids represent obstacles. At each step, the agent can move to the adjacent top, left, right or bottom grids. A possible path, in red, allows the agent to fulfill its goal, and minimizes the total distance it needs to travel. The environment in Figure~\ref{fig_path_planning} can be represented by a geographical graph $\mathcal{G}=(\mathcal{V},\mathcal{E})$. In this graph, nodes in $\mathcal{V}$ are grids not blocked by obstacles, and edges in $\mathcal{E}$ link adjacent (non-diagonal) grids. Edges are assigned a default weight of $1$ as we suppose adjacent grids to be equidistant from one another.

Next, we describe some popular deterministic heuristic-based algorithms for path-planning related problems. These algorithms are well-suited for planning domains with low dimensionality.

\subsection{A*}
\index{Algorithms! Planning! A*}
The A* algorithm \citep{hart1968formal} is a popular best-first search approach to compute an optimal path. Note that A* and related algorithms remain applicable more broadly in other planning domains than path-planning, which is what makes them so popular. A* can be considered as a specialized form of \textit{Dynamic Programming} (DP) \index{Algorithms! Search! DP} \citep{bellman1966dynamic}. DP essentially breaks down a problem into sub-problems in a recursive fashion and seeks to find the optimal choice to make at each step. This can be expressed as a search tree, which a DP algorithm will explore entirely to return an optimal solution. On the other hand, A* differs in that it will guide search towards most promising states first in order to potentially save a significant amount of computation. In path-planning problems, states are graph nodes and transition cost from a state to another is the cost of the edge linking the corresponding nodes in the graph. A* is complete: it will always find a solution if one exists in a finite search space. Depending on requirements, a heuristic which guarantees to find an optimal solution can be used, or a heuristic which simply aims to find a good solution very efficiently, even if possibly sub-optimal. We describe this process next.


Let $S$ be the finite set of states A* explores, $s_{start}$ the start state where the agent starts and $s_{end}$ the state the agent wants to transition into to satisfy goal requirements. In order to guide search, A* proceeds in a best-first fashion by keeping track, for any state $s$ it explores, of an estimate cost $g(s)$ it took to reach that state from the start state $s_{start}$. Algorithm initialization is as follows: $\forall s \in S$, ~ $g(s) \gets \infty $ and $g(s_{start}) \gets 0$. Additionally, A* uses a heuristic $h$ which estimates the remaining best cost from any state $s$ to the goal state $s_{goal}$. The heuristic can be \textit{admissible} to ensure that A* will return an optimal path: once the goal is reached, the path found is guaranteed to be optimal. The fact that the heuristic is admissible means that for any state $s$, $h(s)$ is lower than or equal to the actual cost of the optimal path from $s$ to $s_{goal}$. A* maintains a priority queue, the $OPEN$ list, in which it inserts states by their $'f'$ value. For any state $s \in S$, $f(s) = g(s) + h(s)$. It then proceeds to extract the state $s_{min}$ with the lowest such value in the $OPEN$ list. A* then develops all neighboring states $s' \in S'\subset S$ it can transition into from the state $s_{min}$. For each of those states, costs are updated if possible. More specifically, $\forall s' \in S'$, if $g(s_{min}) + TC(s_{min}, s') < g(s')$ then $g(s') \gets g(s_{min}) + TC(s_{min}, s')$. Here, $TC(s,s')$ returns the transition cost from state $s$ to state $s'$. Additionally, if $g(s')$ is updated, the state $s'$ is added to the $OPEN$ list with its new $f$ value (or its $f$ value is updated if already present in the $OPEN$ list). The best predecessor state for $s'$ is also stored in memory if $g(s')$ is updated, \ie $prev(s') \gets s_{min}$, where function $prev$ stores a predecessor for each state. The A* algorithm will keep extracting states from the $OPEN$ list until the goal state $s_{goal}$ is extracted, at which point a path has been found (and is optimal if an admissible heuristic is used) from $s_{start}$ to $s_{goal}$. 


\subsection{Incremental Planning}

In some scenarios, the agent might not have accurate information about graph structure. The agent may acquire more accurate information about graph structure only when it has started travelling on a computed plan. This is also the case for autonomous vehicles agents if the explored terrain, represented by a graph, is inaccurate at the time of path-planning. It may also be the case if terrain structure changes are likely to happen frequently. The agent will only be able to take into account corrections as it is exploring the terrain.

If the agent computes a path from $s_{start}$ to $s_{goal}$, proceeds on the path, and observes graph changes along the way  (\eg edge connection or weight modifications), the computed path may turn out to be in fact sub-optimal after taking into account the new graph structure. In order to compute the new optimal path from the agent's current position $s$ on the path (when the change is observed) to $s_{goal}$, two possibilities exist. The agent can re-plan from scratch in order to compute the shortest path from $s$ to $s_{goal}$. This approach can however cause expensive computations that may be avoidable (\eg if the change in the graph does not change the optimality of the shortest path, or if it's a minor change that can be fixed with a small modification). The other possibility is to leverage information from the previously computed shortest path in order to repair it and make it optimal again. This is the approach taken in \textit{incremental} planning algorithms. Two particularly popular such algorithms are the D* algorithm \citep{stentz1995focussed} and an improved lighter version of D*, the D* Lite algorithm \citep{koenig2002improved}. D* Lite is quite efficient and remains a method of choice even now, with a wide array of applications relying on it.

Simply put, D* and D* Lite aim to re-expand and develop only parts of the search space relevant to registered graph changes and the potential new current state the agent is in.
The following provides an overview of how D* Lite operates. First, it computes a path from $s_{start}$ to $s_{goal}$ using backwards A*. Backwards A* works in the same way as A* except the search is done backwards: from the goal state $s_{goal}$ to the start state $s_{start}$. Furthermore, a consistency criterion is used for each state $s$ explored. This criterion compares the cost of the optimal path found from state $s$ to $s_{goal}$ to the minimum of the costs to $s_{goal}$ obtained from each neighboring state plus the transition cost to said neighboring states. The state is said to be consistent if they are equal. Otherwise, it is said to be inconsistent (either overconsistent or underconsistent if respectively higher than or lower than). When a change is observed in the graph while the agent is proceeding on the computed shortest path, edges are updated, and the resulting inconsistent states are re-processed in a defined priority order. Once the process is over, the path has been repaired and is optimal again.

Sometimes changes observed which would result in no impact on the optimality of the path will still require computations by D* and D* Lite to guarantee optimality. Such is the case if some edge weights, all of which are outside the computed path, increase. 
The algorithm would still need to reprocess states becoming inconsistent due to their connection to edge changes before guaranteeing optimality, even though it is clear the path computed is still optimal. To address this issue, a modified version of D*, delayed D* \citep{ferguson2005delayed}, has been proposed. To avoid useless computations in such situations, delayed D* initially ignores underconsistent states and only focuses on overconsistent states first. This enables it to potentially save a lot of computations in such cases, making it more suited than D* Lite in some planning domains \citep{ferguson2005delayed}. \index{Algorithms! Incremental Planning! D*} \index{Algorithms! Incremental Planning! D*Lite} \index{Algorithms! Incremental Planning! Delayed D*}

Another incremental approach worthy of note is Lifelong Planning A*  (LPA*) \index{Algorithms! Incremental Planning! LPA*} \citep{koenig2002incremental}. LPA* starts by running an A* instance to determine an optimal path from a start state $s_{start}$ to a goal state $s_{goal}$. Once edge changes are observed, it uses previous search information to re-compute an optimal path more efficiently in a similar way to D*. The main difference with D* is that LPA* does not allow $s_{start}$ and $s_{goal}$ to be modified. In other words, the approach can only be used before the agent starts moving on the path, in case some last-minute changes are learned (presumably remotely). It is thus unsuitable in situations where the agent observes changes as it is already moving on a computed path and needs to adjust the plan from a new position.
More recently, \citep{przybylski2017d} proposed the D* Extra Lite \index{Algorithms! Incremental Planning! D* Extra Light} algorithm. Similarly to D* Lite, D* Extra Lite is based on A* and propagates changes to the previously processed search space in order to re-optimize a path. Unlike D* Lite, the reinitialization of the affected search space is achieved by cutting search tree branches.
This allows the algorithm to often outperform D* Lite, with experiments suggesting it can be almost up to twice faster on typical path-planning problems.

Previously described approaches are applicable in graphs, and are therefore well-suited to, for example, grid environments where agents can move with 45 or 90 degree angles. Such a representation of the environment can cause the optimal path in the graph to actually be sub-optimal in reality. In \textit{any angle} path-planning, the agent can take any angle to move around in its environment. Some incremental planning work have also emerged for such environments. \citep{ferguson2007field} introduce Field D*, \index{Algorithms! Incremental Planning! Field D*} an adaptation of the D* algorithm for any angle path-planning, which reportedly returns a solution path often close to the optimal solution. Other works include Theta* \index{Algorithms! Planning! Theta*} from \citep{nash2007theta}. Based on A*, Theta* is shown to give even shorter paths than Field D*, though not necessarily optimal either. However, Theta* lacks Field D*'s fast replanning capabilities. Finally, \citep{harabor2016optimal} introduced ANYA, \index{Algorithms! Incremental Planning! ANYA} which they show to be significantly faster than previous approaches. Moreover, ANYA also guarantees to find optimal any-angle paths. 

\subsection{Anytime Planning}
In some situations a path needs to be computed quickly. Such could be the case for example for an agent detecting possible obstruction on a planned path while in movement. A solution would be required by the agent as fast as possible to avoid having to come to a complete stop and waste time while re-computing a path. Computing a new optimal path can quickly become very hard, 
even for incremental algorithms if the number of search states required to be re-processed is high. In such a situation, it can be acceptable to compute a solution which is not guaranteed to be optimal very quickly first, so that the agent can keep moving. In the remaining time available (\eg time for the agent to reach decisive points), the previously computed (likely sub-optimal) path can be improved. \textit{Anytime} algorithms, sometimes referred to as \textit{Hierarchical} path-planners,  are designed to address that problem. They build a likely sub-optimal path very quickly and improve the path in the remaining time available. 

There have been various works on anytime algorithms. \citep{likhachev2003ara} introduced the well-known Anytime Repairing A* (ARA*). \index{Algorithms! Anytime Planning! ARA*} This algorithm is made of successive weighted A* searches. In a weighted A* search, the heuristic function $h$ used is multiplied by a factor $\epsilon  > 1$. In doing so, substantial speedup is often provided at the cost of solution optimality. ARA* executes successive weighted A* searches with a decreasing inflation factor $\epsilon$, each of which uses information from previous searches and provides a sub-optimality bound. During each weighted A* search, ARA* considers only states whose costs at the previous search may not be valid anymore due to the new, lower $\epsilon$ value. Another anytime algorithm is the Anytime Weighted A* (AWA*) \index{Algorithms! Anytime Planning! AWA*} \citep{hansen2007anytime}, which is very similar to ARA*. Authors show that AWA* is seven times faster than ARA* on certain domains such as the sliding-tile planning problem of eight puzzles.

From another perspective, \citep{likhachev2005anytime} introduced Anytime Dynamic A* (AD*). \index{Algorithms! Anytime Planning! AD*} Unlike previous approaches, AD* does not differentiate incremental and anytime approaches. Instead, it provides a framework which combines the benefits of both to provide solutions efficiently to hard dynamic problems. Experiments are carried out in an environment where a robotic arm is manipulating an end-effector through a dynamic environment and show AD* generating significantly better trajectories than ARA* and D* Lite in the same time budget. \citep{botea2004near} presented Hierarchical Path-Finding A* (HPA*). HPA* \index{Algorithms! Anytime Planning! HPA*}
proceeds to divide the environment into square clusters with connections, making an abstract search graph which is searched to find a shortest path. Another approach, Partial-Refinement A* (PRA*) \index{Algorithms! Anytime Planning! PRA*} \citep{sturtevant2005partial}, builds cliques of nodes to construct a multi-level search space. The original problem is reduced to finding a set of nodes on the optimal shortest path. However, both HPA* and PRA* address homogenous agents in homogenous-terrain environments. An extension of HPA*, Annotated Hierarchical A* (AHA*), \index{Algorithms! Anytime Planning! AHA*} has been proposed by \citep{harabor2008hierarchical}. It is still one of the most advanced anytime path-planning algorithms to date. AHA* is able to deal with heterogeneous multi-terrain environments by reducing them to simpler single-size, single-terrain search problems. Authors' experiments suggest that near-optimal solutions are returned by the algorithm for problems in a wide range of environments, with an exponentially lower search effort than A*.

\subsection{Probabilistic Methods for Path-Planning}
\begin{figure}[tbh]
\centering
\includegraphics[scale=0.3]{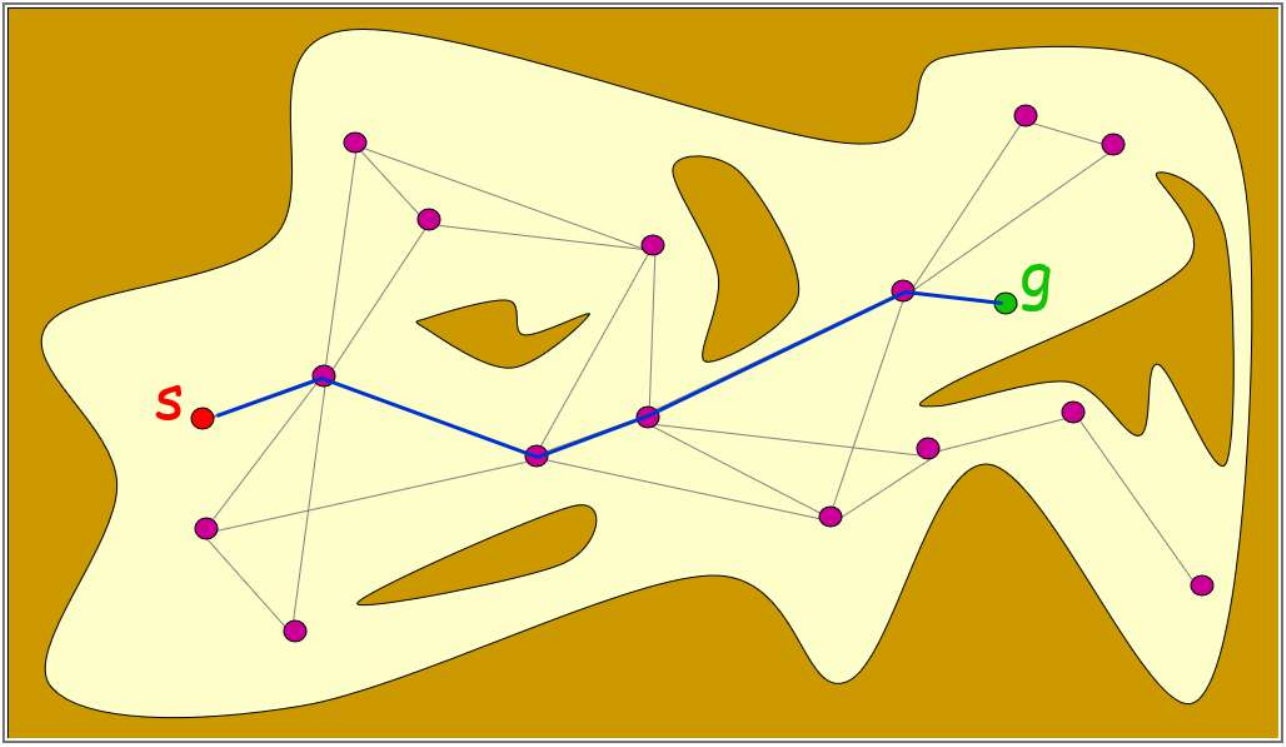}
\caption[A Probabilistic roadmap]{\textbf{A Probabilistic roadmap.} The white space represents feasible states, purple points milestones. Point 's' is the start state, point 'g' the goal state. The shortest path on the roadmap linking s to g is shown in blue. Credit for the picture goes to Jean-Claude Latombe. Source: \citep{probroadmap}}
\label{fig_prm}
\end{figure}

In high-dimensional search spaces, probabilistic approaches can provide a solution quickly but not necessarily an optimal one. We describe two popular approaches, Probabilistic Roadmaps (PRM) \index{Algorithms! Probabilistic Methods! PRM} \citep{kavraki1996probabilistic} and Rapidly-exploring Random Trees (RRT)
\citep{lavalle1998rapidly}. The intuition behind PRMs is to generate random 'points' in the search space, connect these points to nearby points, and repeat the procedure until a path can be computed from the start state $s_{start}$ to the goal state $s_{goal}$ by moving along these points. More specifically, PRM starts by generating random states. It checks whether the generated states are valid, \ie if they do not possess contradictory features (\eg for the robot manipulator in \ref{fig_motion_planning}, one would need to check if the combination of angles does not leave the robot arm in an impossible position). Invalid states are removed, and remaining states are named "milestones". Each milestone is connected to its $k$-nearest neighbor states, $k$ being a parameter. The process is repeated until the roadmap (the milestones and their connections) becomes dense enough and a connection between $s_{start}$ and $s_{goal}$ is created. A shortest path on the roadmap is then computed between  $s_{start}$ and $s_{goal}$. PRM is \textit{probabilistically complete}, \ie as the roadmap building process goes on in time, the probability that the algorithm will find an existing path from $s_{start}$ to $s_{goal}$ tends to 1. Figure ~\ref{fig_prm} illustrates a PRM.  Notable follow-up works include Hierarchical PRMs \citep{collins2003hprm}, which are a variant of PRMs refined recursively, providing better performance at finding narrow passages than uniform sampling. Other works have attempted to improve the efficiency of PRMs by altering the state sample generation process. Recently, \citep{kannan2016robot} built a PRM variant with adaptive sampling. They assign probabilities to different samplers dynamically based on the environment and use the one with the highest probability. \citep{ichter2020learned} proposed to learn to identify 'critical' states with a neural network from local environment features, \ie states that are key to building the wanted path (\eg doorways in an office environment). They draw these critical samples more often and thus are able to build a hierarchical roadmap more efficiently, with reportedly up to three order of magnitude improvements in computation time.

An RRT, \index{Algorithms! Probabilistic Methods! RRT} on the other hand, starts growing a tree rapidly from the start state $s_{start}$, which is considered to be its root. To do so, RRT repeatedly uses a randomly sampled state $s_{rand}$ and attempts to connect $s_{rand}$ to any nearest state in the tree via feasible paths. When successful, RRT expands the size of the tree further with the addition of $s_{rand}$ and intermediary states found on the path. The sampling of random states is done in a way which expands the tree towards unsearched areas of the search space. Furthermore, for a randomly generated state $s_{rand}$ the length of its connection to the tree is limited by a growth factor. If the total length of the connection is above this distance, $s_{rand}$ is dropped and $s_{rand}'$, the state at the maximally allowed distance from the tree along the connection is selected instead. In this manner, the position of the randomly generated samples determines towards which areas the tree gets expanded, while the growth factor limits how far the tree is expanded in those directions. A drawback of RRTs is that they tend to often converge to non-optimal solutions. To address this issue,
\citep{karaman2010incremental} introduced RRT*, \index{Algorithms! Probabilistic Methods! RRT*} which they showed to almost surely converge towards the optimal path without any significant overhead against RRT. We describe some notable follow-up works which are variants of RRT*. \citep{adiyatov2013rapidly} proposed a variant called RRT* Fixed Nodes (RRT*FN). \index{Algorithms! Probabilistic Methods! RRT*FN} Since there is no limit to the number of nodes RRT* can develop, the algorithm is not suited for embedded systems with limited memory. RRT*FN aims to solve the issue by using a node removal procedure which allows it to limit the number of nodes developed without hindering the convergence of the algorithm towards an optimal solution. \citep{gammell2014informed} proposed informed-RRT*, \index{Algorithms! Probabilistic Methods! Informed-RRT*} a variant which uses a heuristic to shrink the planning problem to subsets of the original domain. Informed-RRT* reportedly outperforms RRT* in rate of convergence, final solution cost, and ability to find difficult passages. More recently, \citep{lai2019balancing} presented Rapidly-exploring Random disjointed-Trees* (RRdT). \index{Algorithms! Probabilistic Methods! RRdT} It is a RRT* variant which explores the search space with locally exploring disjointed trees and actively balances global exploration and local-connectivity exploitation. This is done by expressing the problem as a multi-armed bandit problem, and leads to improved performance.


\section{Graph Representation Learning with Graph Neural Networks}
\subsection{Neural Networks}
We start by giving a brief description of neural networks and  convolutional neural networks, architecture types from which graph neural networks originated.

Neural Networks (NNs) \index{Neural Networks! MLP} allow abstraction of data by using models with trainable parameters coupled with non-linear transformations of the input data. In spite of the complex structure of a NN, the main mechanism is straightforward. A \emph{feedforward neural network}, or \emph{Multi-Layer Perceptron (MLP)}, with $L$ layers describes a function $f_{\thetab}(\xb) = f(\xb; \thetab): \real^{d_{\xb}} \rightarrow \real^{d_{\hat{\yb}}}$ that maps an input vector $\xb \in \real^{d_{\xb}}$ to an output vector $\hat{\yb} \in \real^{d_{\hat{\yb}}}$. Vector $\xb$ is the input data that we need to analyze ($\eg$ an image, a signal, a graph, etc.), while $\hat{\yb}$ is the expected decision from the NN ($\eg$ a class index, a heatmap, etc.). The function $f$ performs $L$ successive operations over the input $\xb$:
\begin{align}
  h^{(l)} = f^{(l)}(h^{(l-1)}; \theta^{(l)}), \qquad l=1,\dots,L
\label{intro_eq:layers}
\end{align}
where $h^{(l)}$ is the hidden state of the network ($\ie$ features from intermediate layers, corresponding to intermediary values) and  $f^{(l)}(h^{(l-1)}; \theta^{(l)}): \real^{d_{l-1}} \mapsto \real^{d_{l}}$ is the mapping function performed at layer $l$; $h_0=\xb$. In other words: $$f(\xb)=f^{(L)}(f^{(L-1)}(\dots f^{(1)}(\xb)\dots))$$
\noindent Each intermediate mapping depends on the output of the previous layer and on a set of trainable parameters $\theta^{(l)}$. We denote by ${\thetab=\{\theta^{(1)},\dots,\theta^{(L)}\}}$ the entire set of parameters of the network.
The intermediate functions $f^{(l)}(h^{(l-1)}; \theta^{(l)})$ have the form:
\begin{align}
f^{(l)}(h^{(l-1)}; \theta^{(l)}) = \sigma\left( \theta^{(l)} h^{(l-1)} + b^{(l)} \right) , 
\label{intro_eq:linear}
\end{align}
where $\theta^{(l)}\in\real^{d_l\times d_{l-1}}$ and $b^{(l)}\in\real^{d_l}$ are the trainable parameters and the bias, while $\sigma(\cdot)$ is an \emph{activation} function, \ie a function which is applied individually to each element of its input vector to introduce non-linearities.
Intermediate layers are actually a combination of linear classifiers followed by a piecewise non-linearity. Layers with this form are termed \emph{fully-connected layers}.

NNs are typically trained using labeled training data from a dataset, $\ie$ a set of input-output  pairs $(\xb_i, \yb_i)$,  $i=1,\dots,N$, where $N$ is the size of the dataset. During training we aim to minimize the training loss:
\begin{align}
\mathcal{L}(\thetab) = \frac1N\sum_{i=1}^N \ell(\hat{\yb}_i,\yb_i) ,
\label{intro_eq:loss}
\end{align}
where $\hat{\yb_i}=f(\xb_i; \thetab)$ is the estimation of $\yb_i$ by the NN and ${\ell: \real^{d_L}\times \real^{d_L} \mapsto \real}$ is a loss function which measures the distance between the true label $\yb_i$ and the estimated one $\hat{\yb_i}$. Through \emph{backpropagation}, the information from the loss is transmitted to all $\thetab$ and gradients of each $\theta_l$ are computed w.r.t. the loss. The optimal values of the parameters $\thetab$ are then searched for via Stochastic Gradient Descent (SGD) which updates $\thetab$ iteratively towards the minimization of $\mathcal{L}$. The input data is randomly grouped into mini-batches and parameters are updated after each pass. The entire dataset is passed through the network multiple times and the parameters are updated after each pass until reaching a satisfactory optimum. 
In this manner all the parameters of the NN are learned jointly and the pipeline allows the network to learn to extract features and to learn other more abstract features on top of the representations from lower layers.

In recent years, NNs, in particular Deep Neural Networks (DNNs), have achieved major breakthroughs in various areas. While the fundamental principles of training neural networks are known since many years, the recent improvements are due to a mix of availability of large datasets, advances in GPU-based computation and increased shared community effort. Similarly to NNs, DNNs enable a high number of levels of abstraction of data by using models with millions of trainable parameters coupled with non-linear transformations of the input data. It is known that a sufficiently large neural network can approximate any continuous function \citep{funahashi_1989}, although the cost of training such a network can be prohibitive.

\begin{figure}[tbh]
\centering
\includegraphics[scale=0.4]{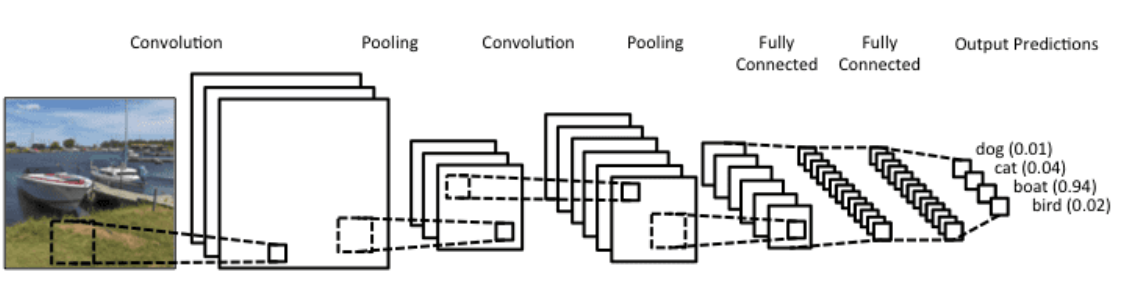}
\caption[A convolutional neural network] {\textbf{A convolutional neural network.} The input of this CNN is an image, the output a prediction of a class among the following set of classes: \{dog,cat,boat,bird\}.}
\label{intro_fig_fig_cnn}
\end{figure}

Convolutional Neural Networks (CNNs) \index{Convolutional Neural Networks! CNN} ~\citep{fukushima_1982, lecun_1995} are a generalization of multi-layer perceptrons for 2D data. In convolutional layers, groups of parameters (which can be seen as small fully-connected layers) are slided across an input vector similarly to filters in image processing. This reduces significantly the number of parameters of the network since they are now \textit{shared} across locations, whereas in fully connected layers there is a parameter for element of the input. Since the convolutional units act locally, the input to the network can have a variable size. A convolutional layer is also a combination of linear classifiers (equation \ref{intro_eq:linear}) and the output of such layer is 2D and is called \textit{feature map}. CNNs are highly popular in most recent approaches for computer vision problems. Figure~\ref{intro_fig_fig_cnn} shows a CNN architecture.

\subsection{Graph Neural Networks}

In this section, we discuss various architectures of Graph Neural Networks (GNNs). GNNs are generalizations of CNNs to non-Euclidean data and aim to learn graph representations. Although no common groups have been precisely defined for GNNs, they tend to belong to four categories:

\begin{itemize}

    \item \textbf{Converging Recurrent Graph Neural Networks (CRGNN)}. These architectures of neural networks mostly include the first works on extending NNs to graphs.
    
    \item \textbf{Graph Convolutional Networks (GCN)}. These networks are mostly inspired from the application of CNNs to graphs and are well-suited to supervised-learning for node classification.

    \item \textbf{Recurrent Graph Neural Networks (RGNN)}. RGNNs are designed to process input graphs for which a temporal sequence ordering exists. They should not be confused with CRGNNs, which are 'recurrent' in the sense they apply a process on the input graph repeatedly until convergence.

\end{itemize}

These networks have different processing architectures but have the following common point. They take as input a graph in which nodes, and possibly edges, have features. They use intermediary layers, each of which produces new features for nodes (and possibly edges, depending on the architecture type). Let $\mathcal{G=(V,E)}$ be an input graph of these network architectures with its set of nodes $\mathcal{V}=(v_1,v_2,...,v_n)$ and its set of edges $\mathcal{E}$. We denote as $A$ the adjacency matrix of $\mathcal{G}$, $X= (x_{v_1}, x_{v_2}, ..., x_{v_n})$ the matrix of node feature vectors of the input graph $\mathcal{G}$ and $H^{(l)}=(h_{v_1}^{(l)}, h_{v_2}^{(l)}, ..., h_{v_n}^{(l)})$ the matrix of node feature vectors after the input graph $\mathcal{G}$ has been processed by $l$ layers. Each $x_{v_i}$  is the feature vector of node $v_i$ of the input graph $\mathcal{G}$ and each $h_{v_i}^{(l)}$ the feature vector of node $v_i$ after the $l^{th}$ layer. Finally, as the layer architecture types we describe next apply a similar process to each graph node, the same layer can be used on input graphs with any number of nodes, although the number of features per node needs to be fixed. In other words, a same GNN with layers made of these architectures can process input graphs with any number of node.

\subsubsection{Converging Recurrent Graph Neural Networks}

The idea behind CRGNNs was initially introduced in \citep{sperduti1997supervised}, with a contribution termed \textit{generalized recursive neuron}, extending the idea of applying neural networks to inputs with structures. Those structures were essentially limited to
acyclic graphs because of computational constraints at the time. In follow-up works, \citep{scarselli2008graph} extend this with an architecture capable of processing acyclic, cyclic, directed and undirected graphs. To that end, neighborhood information among graph nodes is exchanged repeatedly until convergence. The following formula describes how information is updated from layer $l$ to layer $l+1$ for node $v$: 

\begin{equation}
\label{intro_eq_CRGNN}
h_{v}^{(l)} = \sum_{w \in N(v)} f(x_{v}, x_{e_{vw}}, x_w, h_w^{(l-1)})
\end{equation}

\noindent where: 
\begin{itemize}
    \item $h_v^{(l)}$ and $h_v^{(l-1)}$ respectively designate the vector feature of node $v$  after layer $l$ and layer $l-1$; $h_v^{(0)} = x_v$.
    \item $N(v)$ designates the nodes connected to node $v$ with an edge.
    \item $x_{e_{vw}}$ is the feature vector of the edge connecting $v$ and $w$.
    \item $f$ is a parametric function, called \textit{local transition function} by Scarselli \etal
\end{itemize}

Intuitively, the information update of node $v$ from a layer $l-1$ to a layer $l$ proceeds in the following manner for each node. For each neighboring node $w$, a parametric function $f$ takes as input the following elements: the input feature vectors of node $v$, edge $(v,w)$, node $w$, as well as the feature vector of node $w$ after layer $l-1$. The sum of the output of $f$ for each neighbor of $v$ makes the new feature vector of node $v$ after layer $l$. Moreover, to ensure convergence after applying layers repeatedly, function $f$ needs to be a \textit{contraction map} which reduces the distance between inputs and satisfies this property: $$ \forall z \in \real^{m} ~~ \exists \mu \in ]0,1[ ~~ s.t ~~ \forall (x,y) \in \real^{m' \times m''} : ~~  \| f(x,z) - f(y,z) \|  \leq \mu \|x-y\|$$ where $\| \cdot \|$ denotes a vectorial norm.

A convergence criterion also needs to be defined. Layers are applied recursively on each node in parallel until this criterion is satisfied. The converged node feature vectors $h_{v_i}^{*}$ of each node $v_i$ can then be forwarded to an output layer to perform either node classification tasks, edge classification tasks (by using for example a MLP which takes as input converged features $h_{v_i}^{*}$ and $h_{v_j}^{*}$ and outputs a value for edge $(v_i,v_j)$) or graph-level predictions (\eg  predict a class among a portfolio of classes for the input graph, which is typically done by using one or multiple \textit{pooling} operations such as $max$ or $min$ to reduce the size of the converged graph into a fixed size, enabling the use, for example, of a fully-connected output layer).

A notable issue with this CRGNN architecture is the number of layers which need to be applied to meet the convergence criterion and the possibly ensuing complexity. More recently, a framework was proposed in \citep{li2015gated} to address this issue based on gated recurrent units \citep{cho2014learning}. This allows Li \etal~to only require a fixed number of layers to process an input graph, thereby lifting the constraints associated with the convergence criterion. Nevertheless, the approach in \citep{li2015gated} requires Back-Propagation Through Time (BPTT) to compute gradients when using the model in a loss function, which can cause severe overhead.

CRGNNs are mostly pioneer works which inspired the next architectures we describe, and even the newest CRGNN approach presents computational issues due to BPTT.

\subsubsection{Graph Convolutional Networks}
Unlike CRGNNs where a fixed recurrent model is applied repeatedly, GCNs use a fixed number of graph convolutional layers, each of which is different and has its own set of trainable parameters. GCNs are inspired from CNNs. They generalize their operations from grid-structured data (images) to graph-structured data. There are two main categories of GCNs: \textit{spectral-based} and \textit{spatial-based}. Spectral-based approaches use signal processing to define the neighborhood of a node and the ensuing feature update process, while spatial-based approaches rely directly on spatially close neighbors in the graph.

\paragraph{Spectral-based GCNs}

Spectral-based architectures use the spectral representation of graphs and are thus limited to undirected graphs. They were  introduced in \citep{bruna2014}. The following layer propagation rule is used to compute $H^{(l)}$, the matrix of all node feature vectors at layer $l$, from $H^{(l-1)}$:

\begin{equation}
\label{intro_eq_bruna}
H^{(l)} = \sigma (U g_{\theta}(\Lambda) U^{T} H^{(l-1)})
\end{equation}

Here, $U$ denotes the eigenvectors of the normalized graph Laplacian matrix $L = I_N - D^{- \frac{1}{2}} A D^{- \frac{1}{2}}$ ($A$ being the adjacency matrix, $D$ the node degree matrix) and $\Lambda$ its eigenvalues. Function $g_{\theta}(\Lambda)= diag_{\theta}(\Lambda)$ is a filter applied on the eigenvalues with a set of parameters $\theta$. Lastly, $\sigma$ is an activation function. A problem with this approach is that it results in non-spatially localized filters, making it unable to extract local features independently of graph size. In a follow-up work, \citep{defferrard2016convolutional} introduce \textit{ChebNet}. The filters proposed in ChebNet are localized in space. Their idea is to replace $g_{\theta}(\Lambda)$ with a truncation of Chebyshev polynomials $T_k$ of the eigenvalues $\Lambda$: $g_{\theta}(\Lambda) = \sum_{k=0}^{K}\theta_k T_k(\Tilde{\Lambda}) $ where $\Tilde{\Lambda} = \frac{2 \Lambda}{\lambda_{max}} - I_n $, $\theta_k \in \real^{K}$ and $\lambda_{max}$ denotes the highest eigenvalue. Chebyshev polynomials are recursively defined by: $T_k(x) = 2 x T_{k - 1}(x) - T_{k - 2}(x)$; $T_0(x) = 1$; $T_1(x) = x$. After further simplifications, the layer propagation rule is simplified to:
\begin{equation}
    \label{intro_eq_chebnet}
    H^{(l)} = \sigma (\sum_{k=0}^{K} \theta_k T_k(\Tilde{L}) H^{(l-1)})
\end{equation}

\index{Graph Neural Networks! Spectral-based! ChebNet}

\noindent where $\Tilde{L}=\frac{2 L}{\lambda_{max}}- I_n$. More recently, in \citep{kipf_2017}, authors introduce \textit{GCN} by applying a first-order approximation of ChebNet ($K=1$, and $\lambda_{max} = 2$). This enables them to avoid overfitting local neighborhood structures on graphs with unbalanced node degree distributions. Equation \ref{intro_eq_chebnet} becomes:

\index{Graph Neural Networks! Spectral-based! GCN}

\begin{equation}
    \label{intro_eq_gcn_unsimp}
    H^{(l)} = \sigma (\theta_0 H^{(l-1)} - \theta_1 D^{- \frac{1}{2}} A D^{- \frac{1}{2}} H^{(l-1)})
\end{equation}

An additional assumption is made in GCN that $\theta = \theta_0 = - \theta_1$ to further reduce overfitting, and the equation becomes:

\begin{equation}
    \label{intro_eq_gcn_simp_onelayer}
    H^{(l)} = \sigma (\theta ( I_n + D^{- \frac{1}{2}} A D^{- \frac{1}{2}}) H^{(l-1)} )
\end{equation}

Finally, a \textit{re-normalization trick} is used to avoid numerical instabilities such as exploding or vanishing gradients: $I_n + D^{- \frac{1}{2}} A D^{- \frac{1}{2}} \xrightarrow{} \Tilde{D}^{- \frac{1}{2}} \Tilde{A} \Tilde{D}^{- \frac{1}{2}}$, with $\Tilde{A} = A + I_n$ and $\Tilde{D}$ being the degree matrix of $\Tilde{A}$. Kipf and Welling generalize this definition to an input $H^{(l-1)} \in \real^{N \times C}$ where $C$ is the number of features per node at layer $l-1$, $N$ the number of nodes in the input graph. Moreover, They use a weight matrix $W \in \real^{C \times F}$, where $F$ is the desired number of features per node after the layer has been applied. The equation becomes:

\begin{equation}
    \label{intro_eq_gcn}
    H^{(l)} = \sigma (\Tilde{D}^{- \frac{1}{2}} \Tilde{A} \Tilde{D}^{- \frac{1}{2}} H^{(l-1)} W )
\end{equation}

\noindent On a side note, during the information update for a node $v$, GCN takes a weighted sum of vector features from neighbors, where the weight for a neighbor $w$ is given by: $\frac{1}{\sqrt{deg(v) \times deg(w)}}$, where $deg(v)$ refers to the degree of node $v$. Several linear combinations are then applied, to create as many output features as needed for $v$ in the next layer.

\begin{figure}[tb]
\centering
\includegraphics[scale=0.5]{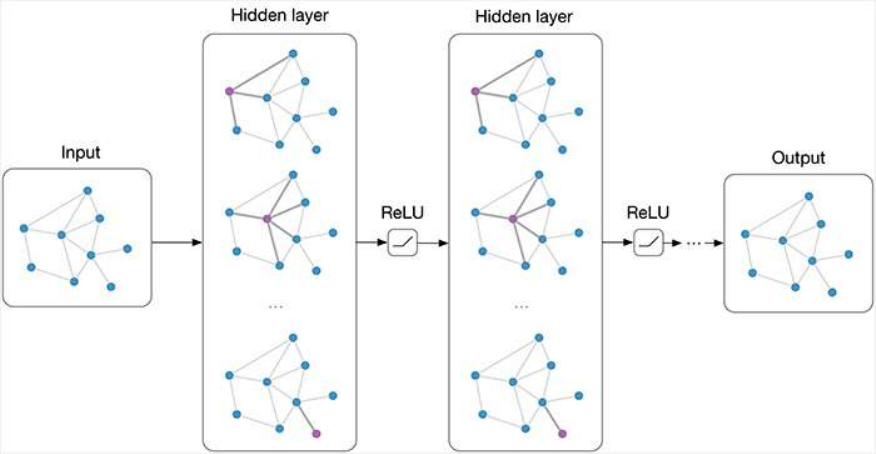}
\caption[A graph convolutional network] {\textbf{A graph convolutional network.} In this illustration, an input graph with node features (and possibly edge features) is processed through multiple graph convolutional layers and $ReLU(\cdot)=max(0,\cdot)$ nonlinearities. An output graph is returned with new, updated node features. Credit goes to Thomas Kipf for the illustration, source: \citep{gcnkipf}}
\label{intro_fig_gcn}
\end{figure}

Lastly, methods presented thus far rely on the adjacency matrix to define relations between nodes, possibly missing on implicit information between nodes. Authors in \citep{li2018adaptive} propose Adaptive Graph Convolutional Network (AGCN) to address this issue. AGCN basically learns a \textit{residual} graph adjacency matrix by learning a distance function which takes as input the features of two different nodes in the graph, enabling it to better capture implicit dependencies.

\index{Graph Neural Networks! Spectral-based! AGCN}

\paragraph{Spatial-based GCNs}
\label{intro_GNN_spatial}
Spatial-based approaches rely on spatially close neighbors to define the feature update step for a node. In this sense, spatial-based GCNs are somewhat similar to CRGNNs in that they propagate node information through edges, although they do not retain the idea of convergence and they stack multiple different layers with different trainable weights. A significant advantage of spatial-based GCNs over spectral-based GCNs is that they can be used on directed graphs.

Among early spatial-based architectures, \citep{micheli2009neural} introduces neural network for graphs (NN4G). \index{Graph Neural Networks! Spatial-based! NN4G} In the NN4G architecture, graph convolutions are performed at each layer (each of which has its own trainable weights). Each convolution basically consists in the sum, for each node, of the feature vectors of neighboring nodes. In this sense, it is somewhat similar to the GCN architecture of \citep{kipf_2017} which performs a weighted sum based on the spectral graph instead. Additionally, NN4G applies residual skip connections between each layer to 'memorize' information. Each new layer is basically linked not only to the previous one, but also to all preceding layers and the input. The following equation defines NN4G's propagation rule, where $\Theta^{(l)}$ and $W^{(k)}$ are weight matrices:

\begin{equation}
    \label{intro_eq_nn4g}
    H^{(l)} = \sigma ( X \Theta^{(l)} + \sum_{k=1}^{l-1} A H^{(k)} W^{(k)} )
\end{equation}

The  Diffusion Convolutional Neural Networks (DCNNs) \index{Graph Neural Networks! Spatial-based! DCNN} proposed in \citep{atwood2016diffusion} brings the concept of diffusion to graphs convolutions. A transition probability is defined when information from a node is passed to a neighboring node, causing the passing of information to converge after applying the process repeatedly. Transition matrices are used to define the neighborhood for a node. The propagation rule for DCNN is:

\begin{equation}
    \label{intro_eq_dcnn}
    H^{(l)} = \sigma ( W^{(l)} \odot P^l H^{(l-1)} )
\end{equation}

\noindent where $W^{(l)}$ is a weight matrix, $\odot$ denotes the element-wise product, $P^l$ (not to be confused with $P^{(l)}$) is $P$ to the power of $l$, with $P = D^{-1}A$ the probability transition matrix.

Message Passing Neural Networks (MPNN), \index{Graph Neural Networks! Spatial-based! MPNN} on the other hand, are a general framework presented in \citep{gilmer2017neural} which aim to regroup different categories of previous works into one single architecture. In MPNNs, during the convolution phase of an input graph, messages are passed between nodes along edges by following an aggregation phase, called \textit{message passing} phase, after which node features get updated in a \textit{message update} phase. Each node $v$ has its feature vector $h_v^{(l-1)}$ updated to $h_v^{(l)}$ based on a message $m_v^{(l)}$:

\begin{equation}
    \label{intro_eq_mpnn_mess_pass}
    m_v^{(l)} = \sum_{w \in N(v)} M_{l-1}(h_v^{(l-1)}, h_w^{(l-1)}, x_{e_{vw}})  
\end{equation}

\begin{equation}
    \label{intro_eq_mpnn_mess_update}
    h_v^{(l)} = U_{l-1}(h_v^{(l-1)}, m_v^{(l)})     
\end{equation}

\noindent where $N(v)$ designates the neighborhood of node $v$; $M_{l-1}$ and $U_{l-1}$ are learned differentiable functions; $x_{e_{vw}}$ is the vector feature of the edge connecting $v$ and $w$. 

A \textit{readout phase} is also introduced (after the last message passing layer $l_{max}$ has been applied), in which a readout function $R$ can optionally compute a feature vector $\hat{y}$ for the whole graph (assuming we want to do some other type of classification than node classification, such as graph-level class prediction):

\begin{equation}
    \label{intro_eq_mpnn_mess_output}
    \hat{y} = R(\{h_v^{(l_{max})}| v \in \mathcal{V}\})    
\end{equation}

\noindent $R$ needs to be invariant to the permutation of node states in order for the MPNN to retain invariance to graph isomorphism. Gilmer \etal~proceed to express previous existing GNN architectures in the literature by specifying the corresponding message passing function $M_{l-1}$, message update function  $U_{l-1}$ and readout function $R$. In their own work, they use an architecture in which $M(h_v, h_w, x_{e_{vw}}) = MLP(x_{e_{vw}})h_w$~. Here, MLP is a multi-layer perceptron which takes as input the feature vector of edge $(v,w)$ and outputs a $out_{c} \times in_{c} $ sized-matrix, $out_{c}$ being the number of desired feature per node after applying the message passing layer and $in_{c}$ the number of feature per node of the input graph provided to the layer. Vector  $h_w$ being of size $in_{c} \times 1$, the matrix multiplication results in a $out_{c} \times 1$ sized-matrix, \ie a vector which has the desired number of new features after the message passing layer is applied. The sum of these vectors for the entire neighborhood defines $m_v$. Gilmer \etal~apply this architecture for node classification tasks on a molecular property prediction benchmark and achieve state-of-the-art results.

Other recent relevant works include GraphSAGE \citep{hamilton2017inductive} \index{Graph Neural Networks! Spatial-based! GraphSAGE} and Graph Attention Networks (GATs) \index{Graph Neural Networks! Spatial-based! GAT} \citep{velivckovic2018graph}. GraphSAGE has been conceived to handle graphs where the number of neighbors for nodes can vary greatly from one node to another. Since always taking into account the entire neighborhood can prove inefficient and costly, graphSAGE uses sampling to define neighborhoods and thus keep a fixed number of neighbors for each node. The propagation rule in a graphSAGE convolution is defined by:

\begin{equation}
    \label{intro_eq_graph_sage}
    h_v^{(l)}= \sigma [W^{(l)} AGG_l(\{h_v^{(l-1)} \} \cup \{ h_u^{(l-1)}, \forall u \in N_r(v) \})]
\end{equation}

\noindent where: $N_r(v)$ designates a fixed-size uniform draw from the set $\{ u \in \mathcal{V}: (u,v) \in \mathcal{E}\}$ and $AGG_l$ is an aggregation function invariant to the permutations of node orderings (\eg mean function). GATs, on the other hand, use an \textit{attention} mechanism which defines weights for each connected pair of nodes. Weights are learned by the attention mechanism so as to reflect the importance of each neighbor of a node $v$. The layer propagation rule is defined by:

\begin{equation}
    \label{intro_eq_gat}
    h_v^{(l)}= \sigma (W^{(l)}  \sum_{u \in \{v\} \cup N(v) } \alpha_{uv}^{(l)} h_u^{(l-1)} )
\end{equation}

\noindent where $N(v)$ refers to the neighborhood of $v$, and $\alpha_{uv}^{(l)}$ is the attention weight.

Additionally, GAT can use multi-head attention mechanisms (\ie have multiple attention heads $\alpha_{uv}^{(l)}$,  $\alpha_{uv}^{(l)'}$,  $\alpha_{uv}^{(l)''}$, etc...). This enables the model to learn different attention schemes in parallel per layer, and shows considerable improvement over GraphSAGE on node classification benchmarks.

\subsubsection{Recurrent Graph Neural Networks}
In many applications, graphs can not only present spatial structure, but also hold temporal dependencies. An example is road network traffic, for which the same graph at different time steps is going to represent the current flow of traffic in the network. RGNNs are inspired by Recurrent Neural Networks (RNNs) and aim to process a sequence of temporal graphs, in order for example to make predictions about future states (\eg how traffic is going to be like in future time steps). For most RGNNs, an RNN-like mechanism is used to memorize and leverage temporal information. Nevertheless, some RGNNs use CNNs to capture temporal information instead. We first describe some RNN-based methods and then some CNN-based approaches.

The idea behind RNN-based RGNNs stems from the recurrent units used in RNNs. When a RNN is used on an input at time step $t$, each hidden layer $h^{(l)^t}$ is computed by combining both the input to the layer $h^{(l-1)^t}$, as well as a 'memory' equal to the output of the same layer at time step $t-1$ : $h^{(l)^{t-1}}$.

Main such works include Structural-RNN (S-RNN) \index{Graph Neural Networks! Recurrent! S-RNN} \citep{jain2016structural}. S-RNN uses different RNNs to handle both node and edge information, namely \textit{nodeRNN} and \textit{edgeRNN}. Diffusion Convolutional Recurrent Neural Network (DCRNN) \index{Graph Neural Networks! Recurrent! DCRNN} \citep{li2017diffusion} is an encoder-decoder framework which applies gated recurrent units on the DCNN architecture. In \citep{seo2018structured}, a Long Short-Term Memory (LSTM) network is combined with the ChebNet graph convolution operator. LSTMs are a popular type of RNN architecture because they are able to maintain a longer memory than RNNs.

CNN-based RGNNs, on the other hand, abandon the idea of keeping a memory and instead use a CNN jointly with a graph convolution operator to capture temporal and spatial information at the same time. Their advantage over RNN-based RGNNs is that they do not require backpropagation through time for gradient computation. The idea is that for each node $v$ in the input graph, a 1D-CNN is applied and temporal information from previous states of the node is aggregated. Next, a graph convolutional layer is applied on the aggregated temporal information to aggregate spatial information. This process is repeated for each layer.

Graph WaveNet \citep{wu2019graph} \index{Graph Neural Networks! Recurrent! Graph WaveNet} introduces a framework with a self-adaptive adjacency matrix. This allows Graph WaveNet to learn latent structures, which can help discover implicit temporal dependencies between nodes in the graph. Lastly, \citep{guo2019attention} introduce an Attention based Spatial-Temporal Graph Convolutional Network (ASTGCN) \index{Graph Neural Networks! Recurrent! ASTGCN} to solve traffic flow forecasting problems. ASTGCN builds on STGCN by introducing attention mechanisms both for spatial and temporal aggregation. This allows ASTGCN to outperform state-of-the-art baselines on real-world datasets from the Caltrans performance measurement system.

\section{Reinforcement Learning}
Reinforcement Learning (RL) consists in designing an agent capable of learning through trial and error by interacting with an environment. This section only aims to briefly describe Markov Decision Processes (MDP)  \citep{bellman1957markovian} and RL concepts. We refer the reader to \citep{sutton2018reinforcement} for a complete introduction to RL. We temporarily use the following notations here, not to be confused with notations from the previous section:

\begin{itemize}
    \item Set $S$: a set of states.
    \item Set $A$: a set of actions.
    \item Set $P$: a set of transition probabilities. The probability $P(s'|s , a) = P_a(s,s')$ refers to the probability of transitioning from state $s \in S$ to state $s' \in S$ after taking action $a \in A$.
    \item Function $R$: a reward function. The transition from state $s \in S$ to state $s' \in S$ after taking action $a \in A$ results in an immediate reward $R(s'|s,a) = R_a(s,s')$.
    
\end{itemize}

In MDPs, the environment is fully observable and actions are instantaneous and non-deterministic. Nevertheless $\forall (s,a) \in (S,A), ~ \exists ! P_a(s,s')$. In other words, after taking action $a\in A$ in the state $s \in S$, a given set of probabilities exist for each state $s' \in S$ which defines the likelihood of transitioning into those states. Moreover, an \textit{immediate reward} function $R_a(s,s')$ defines a given reward obtained from transitioning to a state $s' \in S$ after taking the action $a \in A$ in state $s \in S$. The aim for the agent is to devise an optimal \textit{policy} $\pi^*$ which specifies which action $\pi^*(s) \in A$ to take in any state $s$ in order to maximize the total cumulative reward. 

RL can generally be formulated as a 4-tuple $(S,A,P,R)$, representing an agent interacting with the environment in a MDP. \index{Reinforcement Learning! MDP} The agent interacts with the environment by following a policy $\pi$, and the goal is to find an optimal policy $\pi^*$ by trial and error. Two main RL approaches exist: policy gradient \index{Reinforcement Learning! Policy Gradient} optimization and value function \index{Reinforcement Learning! Value Iteration} optimization. In policy gradient approaches, a parameterizable function $f_\theta$ ($\theta$ being parameters) is used to approximate $\pi$ directly. Through interaction with the environment, the agent is able to learn, given its current policy $f_\theta$, which actions are more suited for given states in $S$. Thus, the agent can modify its policy $f_\theta$ to prioritize these actions. A popular choice for the function $f_\theta$ is neural networks, whose number of layers can be chosen according to the assumed complexity of the function approximated. On the other hand, value function optimization learns two different value functions $Q$ and $V$, which define the policy $\pi$ to follow. The state value function $V$ is defined by: $V^{\pi}(s) = \mathbb{E}_{\pi}(\sum_{i=0}^{\infty}\gamma^i r_{i+1} | s_t= s)$, where $s_t$ refers to the state of the agent at the current time step, $s_{t+1}, s_{t+2}, ...$ at future time steps; $r_{i+1}$ refers to the immediate reward received by the agent at time step $t+i+1$; variable $\gamma \in ]0,1]$ is a discount factor and $\mathbb{E}$ denotes the expectation. Intuitively, $V^{\pi}(s)$ corresponds to the expected sum of rewards when starting in state $(s)$ and following policy $\pi$. The action value function is defined by $Q^{\pi}(s,a) = \mathbb{E}_{\pi}(\sum_{i=0}^{\infty}\gamma^i r_{i+1} | s_t= s, a_t = a)$  where $a_t$ refers to the action taken by the agent at the current starting time step. Intuitively, it corresponds to the expected sum of rewards when starting in state $(s)$, taking action $a$ and following policy $\pi$ afterwards. In environments with large and continuous state spaces, these functions are usually approximated using neural networks.

RL methods also belong to two categories: model-based and model-free. Model-based assumes knowledge of the transition probabilities in the MDP environment, while model-free does not. A popular approach for model-based is value iteration, which consists in updating Q-values by taking into account transition probabilities and known knowledge about transition states. Q-learning \index{Reinforcement Learning! Q-Learning} \citep{watkins1992q} is a popular approach for model-free approaches. It follows the idea of 'pulling' a given Q-value toward the result obtained from a simulation with the environment every time with a learning rate, so as to approximate transition probabilities indirectly. Q-learning follows this update scheme: $$Q(s_t,a_t) \gets  Q(s_t,a_t) + \alpha [r_t + \gamma \max_{a_i} Q(s_{t+1}, a_i) - Q(s_t, a_t)  ]$$
\noindent where $\alpha$ is the learning rate; $\delta_t = r_t + \gamma \max_{a_i} Q(s_{t+1}, a_i) - Q(s_t, a_t)$ is called the temporal difference. 

The use of Deep Q Neural Networks (DQN) \index{Reinforcement Learning! DQN} \citep{mnih2013playing,mnih2015human} has allowed RL tasks to achieve human level gameplay on games from the Atari 2600 plaform. More recently, AlphaGo \index{Reinforcement Learning! AlphaGo} \citep{silver_2016}, an AI program conceived to play the game of Go combining deep CNNs and Monte Carlo Tree Search (MCTS), managed to defeat the world champion of Go. AlphaGo uses supervised learning to learn from expert gameplay, and then refines the learned policy with RL (policy gradient) by playing against itself. AlphaGoZero \citep{silver_2017}, a newer version, is only trained with RL and achieves superior gaming performance than AlphaGo. 

Model-based RL algorithms are usually more efficient than model-free ones since they can leverage planning using known environment dynamics. A solution for model-free approaches would be to learn the dynamics from interactions with the environment. Although learning dynamics which are accurate enough for planning has remained a challenge in model-free approaches, the PlaNet approach from \citep{hafner2019learning} \index{Reinforcement Learning! PlaNet} achieves a breakthrough on this subject for image-based domains. PlaNet learns the dynamics model by relying on a sequence of latent states generated by an encoder-decoder architecture, rather than images directly. PlaNet chooses actions purely by planning in this latent space, this allows it to require far lower interaction with the environment to optimize its policy than previous recent approaches in model-free RL.

Lastly, in some situations, the environment may not be fully observable by the agent. This is the case in Partially Observable Markov Decision Processes (POMDP). Agents get sensory information and derive a probability distribution of states they may likely be in, and need to adapt their policy accordingly. Some popular works dealing with planning in POMDP include  \citep{kurniawati2008sarsop} who introduce a point-based POMDP algorithm for motion-planning, \citep{silver2010monte} who propose an MCTS algorithm  for planning in large POMDPs, \citep{somani2013despot} who present a random scenario sampling scheme to alleviate computational limitations and \citep{zhu2017improving} who propose a Deep Recurrent Q-Network to adapt RL tasks in POMDPs.

\section{Path-Planning and Neural Networks}
A*-based algorithms described previously are fast on small planning domains, but take exponentially longer as domain size and complexity grows. Probabilistic approaches such as PRMs and RRTs on the other hand construct a new graph with random sampling to bypass this complexity, but to guarantee consistent solution quality the sampling would need to be exponential again \citep{lavalle2004relationship}. Therefore, the idea of using neural networks for path-planning has long been a problem of interest, although recent advances in machine learning has made it a viable option only recently. We explore a few such works.

\citep{glasius1995neural} is an early work which specifies 
obstacles into a topologically ordered neural map, and uses a neural network to trace a shortest path. The minimum
of a Lyapunov function is used for convergence for neural activity.  \citep{chen2016dynamic}, a more recent work, relies on \textit{Deep Variational Bayes Filtering} (DVBF) \citep{karl2016deep} to embed dynamic movement primitives of a high dimensional humanoid movements in the latent space of a low dimensional variational autoencoder framework. RL has also been used for such purposes. \citep{levine2013guided} present a guided policy search algorithm that uses trajectory optimization to direct policy learning and avoid poor local optima, where policies are approximated by neural networks. This method is successfully applied to learn policies for planar swimming, hopping, walking and simulated 3D humanoid running. \citep{tamar2016value} introduce a neural network to approximate the value iteration algorithm in order to predict outcomes that involve planning-based reasoning. Their use of CNNs limits their approach to path-planning on 2D grids and not motion planning in general. 

Some \textit{imitation learning}-based approaches have also been proposed. Imitation learning consists in having an expert provide demonstrations, in this case of desired trajectories. A neural network can then be used to approximate the behavior of the expert, and hopefully generalize outside of the scope of provided demonstrations. Imitation learning has been successful in several areas involving complex dynamical systems \citep{abbeel2010autonomous, calinon2010learning}. OracleNet, an extension of imitation learning for path-planning has been proposed recently in \citep{bency2019neural}. OracleNet relies on an LSTM to build end-to-end trajectories in an iterative manner. The LSTM needs to be trained on optimal trajectories that span the
entire configuration space of the considered environment before being used. Those optimal trajectories can be computed by algorithms such as A*. Although the proposed approach can be problematic if the framework needs to be quickly used in a newly known environment and no training time is available, OracleNet achieves performance which makes up for it. Paths are generated extremely fast, scaling almost linearly with dimensions reportedly. On a benchmark comprised of a point-mass robot with multiple degrees of freedom, OracleNet is compared to A* and RRT*. It achieves solution quality reportedly rivaling A* and far above RRT*, while its execution time remains far below the other two. In the context of path-planning under constraints, Osanlou \etal~ have combined a GNN with a constraint programming solver and a branch \& bound tree search algorithm, observing in each case a significant improvement in the computation of solution paths, outperforming A$^*$-based domain-tailored heuristics \citep{osanlou2021constrained, osanlou2019optimal, osanlou2021learning}.

\section{Temporal Planning With Uncertainty}
\label{intro_chapter_dtnu}
Scheduling  in  the  presence  of  uncertainty  is  an  area  of  interest in artificial intelligence. In this section, we present necessary notions and work leading up to the Disjunctive Temporal Network with Uncertainty (DTNU).

Temporal Networks \index{Scheduling! STN} \citep{dechter1991temporal} are a common formalism to represent temporal constraints over a set of timepoints (\emph{e.g.} start/end of activities in a scheduling problem). A Simple Temporal Network (STN) $\Gamma$ is defined by a pair:

$$\Gamma = (A,C)$$ 

\noindent Where: 

\begin{itemize}
    \item $A=(a_1, a_2, ..., a_n) \in \real^n$ is a set of $n$ real controllable timepoint variables.
    \item $C$ is a set of \textit{free} constraints, each of which of the form: $a_j - a_i \in [x_k, y_k]$, where $a_i, a_j \in A$; $x_k \in \{-\infty\} \cup \real$; $y_k \in \real \cup \{+\infty\}$.
\end{itemize}

\noindent A solution to STN $\Gamma$ is a complete set of assignments in $\real$ for each $a_i \in A$ which satisfies all constraints in $C$.

The Simple Temporal Networks with Uncertainty (STNUs) \citep{kn:Ts,kn:ViFa} explicitly incorporate qualitative uncertainty into temporal networks. In STNUs, some events are \textit{uncontrollable}. The only controllable aspect is when they start: how long they take to complete, however, is not known. Although the duration for completion is uncertain, it is often known to be within some bounds. These uncontrollable events are represented by a \textit{contingency link}, \ie a  triplet $(a,[x,y],u)$, where $a$ is a controllable timepoint (representing the start of the uncontrollable event), $[x,y]$ is the bounded duration of the uncontrollable event and $u$ is an uncontrollable timepoint which signifies the end of the uncontrollable event. Uncontrollable timepoint $u$ will occur on its own, at earliest $x$ units of time after execution of $a$, $y$ at latest.

Formally, an STNU \index{Scheduling! STNU} $\Gamma$ is defined as : $$\Gamma = (A,U,C,L)$$

\noindent Where:
\begin{itemize}
    \item $A=(a_1, a_2, ..., a_n) \in \real^n$ is a set of real controllable timepoint variables, which can be scheduled at any moment in time.
    
    \item $U=(u_1, u_2, ..., u_q) \in \real^q$ is a set of uncontrollable timepoint variables.
    
    \item Each uncontrollable timepoint $u_j \in U$ is linked to exactly one controllable timepoint $a_i \in A$ by a contingency link $l \in L$ :  $l = (a_i, [x,y], u_j)$
    
    \item $C$ is a set of free constraints of the same form as with STNs, except constraints can also involve uncontrollable timepoints in addition to controllable timepoints.
\end{itemize}

\noindent We refer to timepoints in general (controllable or uncontrollable) as $V = A \cup U$. Different types of \textit{controllability} exist \citep{kn:ViFa}:

\begin{itemize}
    \item \textit{Strong Controllablity} \index{Scheduling! SC} (SC): An STNU $\Gamma = (A,U,C,L)$ is strongly controllable if there exists at least one universal schedule of controllable timepoints $\{a_1 = w_1, a_2 = w_2, ..., a_n = w_n\}$ which satisfies the constraints in $C$ regardless of the values taken by uncontrollable timepoints $U$.
    
    \item  \textit{Weak Controllablity} \index{Scheduling! WC} (WC): An  STNU $\Gamma = (A,U,C,L)$ is weakly controllable if, for every value outcome of uncontrollable timepoints $U$, there is at least one schedule of controllable timepoints $\{a_1 = w_1, a_2 = w_2, ..., a_n = w_n\}$ which satisfies the constraints in $C$.
    
    \item \textit{Dynamic Controllablity} \index{Scheduling! DC} (DC):  An  STNU $\Gamma = (A,U,C,L)$ is dynamically controllable if there is a reactive strategy which guarantees constraints in $C$ will be satisfied if the scheduling strategy is followed by a controller agent, while observing possible occurrences of uncontrollable timepoints and using this knowledge to adapt decisions. It is said the problem is DC if and only if it admits a valid dynamic strategy expressed as a map from partial schedules to Real-Time Execution  Decisions  (RTEDs) \citep{cimatti2016dynamic}. A partial schedule represents the current scheduling state, \ie the set of timepoints that have been scheduled so far and their timing. RTEDs are popular semantics used to express a DC strategy \citep{hunsberger2009fixing}. RTEDs regroup two possible actions: \textbf{(1)} The wait action, \ie  wait for an uncontrollable timepoint to occur. \textbf{(2)} The $(t, \mathcal{X})$ action, \ie if nothing happens before time $t \in \real$, schedule the controllable timepoints in $\mathcal{X}$ at $t$. A strategy is valid if, for every possible occurrence of the uncontrollable timepoints, controllable timepoints get scheduled  in  a  way  that  all  free  constraints  are  satisfied.

\end{itemize}

\noindent Considerable work has resulted in algorithms to determine whether or not an STNU is DC or not \citep{kn:MoMu2,kn:Mofast}, leading to $\mathcal{O}(N^3)$ worst-case DC-checking algorithms, where $N$ is the number of timepoints of the STNU. These DC-checking algorithms also synthesize valid DC strategies executable in $\mathcal{O}(N^3)$ \citep{hunsberger2016efficient, kn:Mofast}. Disjunctive Temporal Networks with Uncertainty (DTNUs) \index{Scheduling! DTNU} generalize STNUs by allowing the presence of disjunctions in the constraints $C$ or contingency links $L$. Formally, each constraint in $C$ is of the form : $\lor_{k=1}^{q} v_{k,j} - v_{k,i} \in [x_{k},y_{k}]$. Furthermore, each contingency link $l \in L$ is of the form : $(a_i, \lor_{k=1}^{q'} [x_{k},y_{k}] ,u_j)$ where $x_{k} \leq y_{k} \leq x_{k+1} \leq y_{k+1} ~ \forall k = 1, 2, ..., q'-1$. All controllability types for STNUs remain available for DTNUs. The introduction of disjunctions inside $C$ and $L$ renders STNU's $\mathcal{O}(N^3)$ DC-checking algorithms unavailable for DTNUs. In fact, the complexity of DC checking for DTNUs is $PSPACE$-complete \citep{kn:BhWi}, making this a highly challenging problem. The difficulty in proving or disproving DC arises from the need to check all possible combinations of disjuncts in order to handle all possible occurrence outcomes of the uncontrollable timepoints. The only known approach for DC-checking and DC strategy generation for DTNUs is based on expressing DTNUs as timed-game automata (TGAs) \citep{cimatti2014sound}. TGAs can then be solved by the UPPAAL-TIGA software \citep{behrmann2007uppaal}. In \citep{cimatti2016dynamic}, authors express DTNUs as TGAs in the same way, but use a pruning procedure based on satisfiability modulo theory and achieve superior results than with UPPAAL-TIGA. Authors in \citep{osanlou2021time, osanlou2022solving} design a tree search algorithm that searches in \textit{Restricted Time-based Dynamic Controllability} (R-TDC), a subspace of DC. They show R-TDC allows higher strategy search efficiency than TGA-based approaches while retaining very high DC coverage, thus almost always finding a strategy when a DC one exists on considered benchmarks. They also note a significant increase in search performance for harder problems owing to a heuristic based on graph neural network they use for search guidance.

\bibliography{aaai23}

\end{document}